\documentclass[conference]{IEEEtran}
\IEEEoverridecommandlockouts
\usepackage{cite}
\usepackage{amsmath,amssymb,amsfonts}
\usepackage{algorithmic}
\usepackage{graphicx}
\usepackage{textcomp}
\usepackage{xcolor}
\usepackage{array}
\usepackage{caption}
\usepackage{booktabs}
\usepackage{multirow} 
\usepackage[left=1.62cm,right=1.62cm,top=1.9cm]{geometry}
\def\BibTeX{{\rm B\kern-.05em{\sc i\kern-.025em b}\kern-.08em
    T\kern-.1667em\lower.7ex\hbox{E}\kern-.125emX}}
\newgeometry{right=0.62in}
\newgeometry{left=1.62cm,right=1.62cm,top=1.9cm}
\begin{document}

\title{Data-driven building energy efficiency prediction using physics-informed neural networks \\
}
\author{\IEEEauthorblockN{Vasilis Michalakopoulos, Sotiris Pelekis, Giorgos Kormpakis, Vagelis Karakolis, Spiros Mouzakitis
\\Dimitris Askounis} \\
\textit{Decision Support Systems Laboratory} \\
\IEEEauthorblockA{\textit{School of Electrical \& Computer Engineering} \\
\textit{National Technical University of Athens}\\
Athens, Greece \\
\{vmichalakopoulos, spelekis, gkorbakis, vkarakolis, smouzakitis, askous\}@epu.ntua.gr}
}

\maketitle

\begin{abstract}

The analytical prediction of building energy performance in residential buildings based on the heat losses of its individual envelope components is a challenging task. It is worth noting that this field is still in its infancy, with relatively limited research conducted in this specific area to date, especially when it comes for data-driven approaches. In this paper we introduce a novel physics-informed neural network model for addressing this problem. Through the employment of unexposed datasets that encompass general building information, audited characteristics, and heating energy consumption, we feed the deep learning model with general building information, while the model's output consists of the structural components and several thermal properties that are in fact the basic elements of an energy performance certificate (EPC). On top of this neural network, a function, based on physics equations, calculates the energy consumption of the building based on heat losses and enhances the loss function of the deep learning model. This methodology is tested on a real case study for 256 buildings located in Riga, Latvia. Our investigation comes up with promising results in terms of prediction accuracy, paving the way for automated, and data-driven energy efficiency performance prediction based on basic properties of the building, contrary to exhaustive energy efficiency audits led by humans, which are the current status quo. 

\end{abstract}


\section{Introduction}
\label{sec:intro}
\subsection{Background}
Climate change is already making a severe impact on our daily lives and posing a threat to all of the planet's ecosystems. Scientists, governments, and institutions vigorously alert that we are in an uncharted area full of uncertainty \cite{sarmas2022meta}. Addressing this problem requires a significant focus on enhancing energy efficiency. The parties of the Paris Agreement agreed to keep global warming well below 2$^{\circ}$C and pursue efforts to limit it to 1.5$^{\circ}$C \cite{Terhaar2022}. In order to attain this objective, the International Renewable Energy Agency (IRENA) recommends a full decarbonization of the energy sector by the year 2050 \cite{irena2015renewable}. In this context, several EU-funded projects \cite{Karakolis202261, Wehrmeister2022TheApplications, Pau2022MATRYCSDomain} and research papers have recently conducted research related to forecasting \cite{Pelekis2022InPerformance, Pelekis2023ADrivers, Pelekis2023DeepTSF:Forecasting}, flexibility optimization and demand response \cite{Pelekis2023TargetedTechniques,michalakopoulos2024machine}, and power-to-gas system's assessment \cite{Pelekis2023, KOUTSANDREAS2023101233}, and climatic modeling \cite{FRILINGOU2023102934, van2023multimodel, KOASIDIS2023205} with the ultimate goal of maximizing energy efficiency.  

In addition, according to researchers, the building sector accounts for around 40\% of the total global energy consumption \cite{lasvaux2010study}. Numerous studies have been undertaken within this particular sector, all aimed at reducing the energy consumption in buildings, while entire building heat loss tests indicate that dwellings can experience 60\% or greater heat loss than expected by design \cite{stafford2012building}. Moreover, as urbanization continues its expansion, there is a growing demand for energy in buildings. That poses a serious issue that needs to be addressed particularly in the residential building sector. According to \cite{AHMAD2018301}, over 50\% of the studies concentrate on predicting the overall building level energy usage, which captures the total performance of the building. That poses a gap in the literature in terms of data-driven prediction of the contribution of each individual component of the envelope to the building's energy efficiency.

\subsection{Related work}
Energy efficiency investments are crucial for reducing energy consumption and mitigating climate change. However, these investments are often perceived as risky due to uncertainties in energy savings and financial returns. In recent years, researchers have proposed several data-driven approaches \textemdash based on machine learning (ML), artificial neural networks (ANN), deep neural networks (DNN) \textemdash aiming at the prediction and the optimization of energy efficiency in buildings and the quantification of the projected benefits by respective energy efficiency investments. In this context, Sarmas et al. \cite{sarmas2022meta} proposed an approach that is based on stacking ensemble classification ML models \textemdash namely k-Nearest Neighbors, Gaussian Naive Bayes, XGBoost, Random Forest and SVM \textemdash accompanied by a 3-phase meta-learning model. The proposed methodology is evaluated using a set of 312 projects that have been completed in Latvia, and the results indicate that the meta-learner outperforms all baseline classifiers in identifying high and medium potential projects and distinguishing low from high potential ones. 

With respect to reinforcement learning, an optimization method based on graph mining has been introduced by Haidar \cite{HAIDAR2023109806}, which combines a passenger behavior prediction model and a selective reinforcement learning method, where error detection is based on sensors that detect occupancy of rooms in real time. The study concludes that with 50\% prediction accuracy, the proposed system can reduce energy savings by an average of 31.9\% varying based on the building type and the occupants’ behavior. Similarly, Homod et al. \cite{HOMOD2023105689} proposed deep clustering multi-agent cooperative reinforcement learning (DCCMARL) as suitable for such a system control, which supports centralized control with agent learning. DCCMARL leverages a hybrid clustering algorithm to deal with complexity and uncertainty, which is a crucial factor in achieving high performance. The results showed that the agents' ability to manipulate the behavior of the smart building could save energy up to 44.5\% compared to conventional methods.

With regard to simulation-based approaches, Long \cite{LONG2023480} introduced a new integrated model for energy-efficient building envelope design in the early stages using ML algorithms and neural networks, in which Gradient Boosting (GB) is identified as the most well-performing algorithm among others. The results demonstrates savings of 7.52\% in cost and 8.48\% in energy, or 21.17\% in cost and 0.4\% in energy, for a case study in Vietnam. Extending the concept of energy efficiency investment beyond the building domain, Kalogirou \cite{KALOGIROU2004383} modeled a solar energy system and then trained an ANN using the results of a small number of simulations to correlate the collector area and storage tank size with the auxiliary energy required by the system. Subsequently, he employed genetic algorithms to estimate the optimal values of these two parameters. His results yield increased life cycle savings of 4.9 and 3.1\% when using subsidized and non-subsidized fuel prices respectively, compared to traditional trial and error methods. Finally, Borge-Diez et. al \cite{BORGEDIEZ2015821} propose an energy model for geothermal source heat pumps in livestock facilities alongside an analysis of a business case towards the reduction of energy costs, the increase of productivity, and the improvement of the financing system of energy service companies. A sensitivity analysis reveals that the internal rate of return (IRR) ranges from 10.25\% to 22.02\%, making the investment attractive across different locations and climate conditions.

\subsection{Contribution}
In this study we propose a data-driven, physics-informed approach for the estimation of energy performance in residential buildings based on general building information alongside measured heating energy consumption. The primary contributions of our research are listed as follows:
\begin{itemize}
  \item As an alternative to simulations \cite{LONG2023480, KALOGIROU2004383, BORGEDIEZ2015821} and physical energy audits conducted by expert engineers, we train our models on a group of datasets including energy audit data, however during inference the neural network only requires as input general building information that can be easily provided by the household owners themselves.  Such an approach sets an innovative starting point for purely data-driven, audit and/or simulation-free EPCs. 
  \item We provide a novel DL approach, that encompasses linear physics equations in its loss function. This enhanced loss function is used for: i) the calculation of the energy consumption of a building, decomposed at its separate envelope components and their respective linear energy loss equations; ii)  the subsequent validation of said energy consumption calculation based on real energy consumption measurements of a domestic heating network (DHN).
  \item Aiming at an end-to-end building energy performance prediction, we go far beyond the mainstream task of energy consumption prediction \cite{AMASYALI20181192} by also predicting the area and thermal transmittance of each sub-component of the building envelope (windows, doors, floors, roofs, basements). As these are crucial ingredients of an EPC, our approach highlights the specific weak points that urge for energy efficiency renovation actions. Therefore, our framework can be used by researchers, building operators, and funding agencies to easily assess building retrofit programs and motivate household owners to proceed to targeted energy efficiency investments.
\end{itemize}

\subsection{Structure of the paper}
The rest of the paper is organized as follows. In Section \ref{sec:intro} the
problem setting and literature review are presented. In Section \ref{sec:datasets}
the unique datasets used for our research are presented . In Section \ref{sec:methodology} the proposed methodology is described. In Section \ref{sec:experiments} the experiments done using our methodology are presented. Finally, limitations, concluding and future work remarks are provided in Section \ref{sec:lmitations} and \ref{sec:conclusions} respectively.

\section{datasets}
\label{sec:datasets}
In the present study, we use data gathered from residential buildings in Riga, Latvia. There are three datasets in total that are essential for this study. At first, to break it down, we use a vast dataset, namely State Land Dataset, that comprises information for 11734 residential buildings in Riga, Latvia. Subsequently, a dataset of 256 already audited buildings is acquainted, holding details and specifications about the envelope and the energy efficient characteristics of the structures. Thus, it can be split in two, the one containing general information about the audited buildings and the second comprising information about structural material and their energy efficiency. Finally, a third dataset (Energy Consumption dataset (2017-2020)) provides the measured monthly energy consumption of a subset of buildings that were connected to a district heating system, from 2017 to 2020. Notably, each of the datasets is a subset of the State Land Dataset. Table \ref{tab:combined} comprises all the features in detail from all the datasets. 

\begin{table*}[htbp]
  \centering
  \caption{Combined features from the datasets}
  \resizebox{\textwidth}{!}{%
    \begin{tabular}{lllll}
      \toprule
      \textbf{Land dataset} & \textbf{Audit dataset} & \textbf{Envelope Components} & \textbf{Energy Consumption (2017-2020)} & \\
      \midrule
      cadastre\_number & cadastre\_number & cadastre\_number & cadastre\_number & \\
      floors & floors & enclosing\_structure & total\_energy\_consumption\_2017 & \\
      latitude\_centroid & length & material & total\_energy\_consumption\_2018 & \\
      longitude\_centroid & width  & energy\_consumption & total\_energy\_consumption\_2019 & \\
      useful\_area & useful\_area & area & total\_energy\_consumption\_2020 & \\
      geometry & Avg\_indoor\_height & structure\_heat\_loss\_coefficient & & \\
      apartments & apartments & type\_of\_heating & & \\
      serie & serie & total\_structure\_heat\_loss\_coefficient & & \\
      total\_area & total\_area & total\_area & \\
      address & air\_exchange\_rate & total\_energy\_consumption & \\
      perimeter & specific\_heat\_gains & & & \\
      building\_type & building\_type & & & \\
      \bottomrule
    \end{tabular}%
  }
  \label{tab:combined}
\end{table*}

In the first column, general information about various buildings, through multiple fields, are presented. The cadastre number is the unique identification assigned to a property and acts as the primary key for every building processed. This field is used as identification of each building during this study. Next, the address field specifies the street address of each building. Furthermore, the geometry field represents the building's geometric shape using the MultiPolygon format with a spatial reference identifier (SRID) \cite{srid} of 3059, which is the identifier specifically for Latvia. Specifically, the perimeter, the coordinates of the centroid and the MultiPolygon formats are acquainted for further analysis using the PostGIS \cite{postgis2018postgis} extension of PostgreSQL \cite{psqldocs}, which are open source tools suitable for manipulating geographical data. Other essential details include the number of floors, total useful area, the number of apartments, and the total area of each building. Furthermore, the building type is an additional feature that depends on the main construction materials and can be classified as heavy or light. Adding to the previous, the unique feature of the use case is named serie. There exist distinct categories of buildings in Riga, with classifications determined by both their construction period and overarching characteristics. Riga is a city where a significant portion of its architectural heritage was created during the same time frame by a select group of architects, resulting in designs that exhibit minimal differences for each time period. This unique scenario has led to the emergence of twelve distinctive building categories that can describe a huge portion of the residential constructions in the city. These categories hold exceptional value as they yield information of significant importance for the predictions, adding an extra feature that further classifies each building.

In the second and third column of  Table \ref{tab:combined}, detailed information related to building energy audits are presented. Note here that all information derived from these columns is on a yearly basis, as for instance the variables related to energy consumption. The cadastre\_number, the serie, the useful and total area, the number of floors, the number of apartments and the building type are explained in the last paragraph. Additionally, the length and width of the audited buildings are present in this column and can be related to the perimeter of the building from the first column. Another feature that might be of use for another study, is the average indoor height of each floor. For the purposes of this paper the average indoor height is not utilized. Furthermore, the air exchange rate and the specific heat gains of the useful area of the building are encompassed in this dataset. Regarding the third column, valuable information about the envelope components and their energy efficiency characteristics are included. The enclosing structure is denoted into five distinct categories, namely Windows, Doors, Walls, Basement/Slab and Roof/Attic, specifying the specific building envelope component categories that contribute to energy losses and consequently energy efficiency perfomance. Notably, for each one of these categories, details are provided on their material, area,  energy consumption, and structure heat loss coefficient. Crucially, the U-value for each component is also included, which stands for thermal conductivity \cite{thermaltransmittance} and ability to resist heat transfer. This value is crucial in assessing the material's performance in insulating structures and it is defined in Eq. \ref{eq:Uvalue}. 

\begin{equation}
    U = \frac{E}{A}  :  [\text{W/(m}^2\text{K)}]
    \label{eq:Uvalue}
\end{equation}

Where:

\begin{align*}
    A & : \text{Area of component} : \text{[m}^2] \\
    E & : \text{Structure heat loss coefficient} : \text{[W/K]} 
\end{align*}

Finally, in the last column of Table \ref{tab:combined}, the average energy consumption for the period between 2017 and 2020 is presented. The energy consumption was given in a monthly resolution and the sum of all the months creates the total energy consumption feature for each year separately.  
As observed, these datasets provide valuable of information crucial for predicting and understanding energy consumption patterns in residential buildings. The incorporation of a wide array of information and measurements, spanning from the characteristics of buildings to intricate details about envelope components, not only enriches the depth of analysis but also makes a substantial contribution to fulfilling the objectives of the study. Particularly, the fragmenting of the envelope to five distinct categories as well as the series feature, are the ones that are separating this study from others in this field.

\section{Methodology}
\label{sec:methodology}

In this section, we introduce a deep neural network enriched with an custom (enhanced) loss function, that integrates innovative physics-informed equations. Figure \ref{fig:DNN} provides a comprehensive overview of our holistic approach. To provide a thorough understanding of our strategy, we delve into the neural network and our method for predicting five distinct envelope components and their characteristics in Subsection \ref{subsec:dnn}. Subsequently, we present the custom loss function in Subsection \ref{subsec:CLF} which connects the physics-informed function with the Neural Network . Finally, we discuss how the predictions are utilized as input to our physics-informed function in Subsection \ref{subsec:calculations}.

\subsection{DNN architecture}
\label{subsec:dnn}
% title
Aiming to address the complexities of our identified problem and propose a solution for a multi-input, and multi-output regression scenario, our focus lies in utilizing datasets pertaining to all three aforementioned datasets. The deep neural network (DNN) architecture employed in this study consists of an input layer, two hidden layers, and an output layer. Each hidden layer contains 256 neurons, contributing to the expressive capacity of the model. The choice of a two-layer architecture with 256 neurons in each hidden layer was guided by the need to capture complex patterns and relationships within the data while avoiding over fitting.

% title
To achieve comprehensive building energy performance prediction, we extend our scope beyond the typical task of energy consumption prediction by predicting the dimensions and thermal properties of individual building envelope sub-components. Those sub-components are five in number, namely: windows, doors, floors, roofs, and basements. Adding to this, we decided it is more efficient to predict the areas and the U values, which denotes the thermal transmittance of a material, of each component of the envelope in order to provide input to the $F_{}$ function which is described in the following subsection. Subsequently, the output of the $F_{}$ function is the energy consumption of the building, being a linear calculation of the predicted values of the model.In order to conclude the inputs of this function we needed to predict two more values, namely specific heat gains and air exchange rate, that are crucial for the calculation of the energy consumption of a building. Finally, the values that are used as validation of the energy consumption produced by $F_{}$, come from the the Energy Consumption dataset described in the previous section. Notably, this unique methodology allows our solution to be effectively harnessed for renovation purposes.

By predicting specific characteristics for each segment of the building envelope, we enhance the granularity of our insights, enabling more targeted and efficient interventions. Thus, the input features could only be the ones that are common between the land and audit datasets in order for these predictions to have actual meaning. Those are eighteen in number, namely, the useful and total area, the number of floors, the number of apartments, the building type and each of the series categories label encoded into twelve distinct features.

\subsection{Enhanced loss function}
\label{subsec:CLF}
In our research, we employ a customized loss function within a DNN, designed to capture the intricate relationship between predicted values and ground truth data. The foundation of this custom loss function resides in the utilization of the Mean Squared Error (MSE), as expressed in Equation \ref{eq:loss}. However, this loss function's uniqueness lies in its two-sided nature, as it combines two distinct components: the first component encapsulates the error between the model's predictions and observed data ($z$), while the second component pertains to the error within the physics-informed knowledge domain ($y$). This hybrid approach infuses our custom loss function with enhanced capability, allowing it to encapsulate both the predictive and physical aspects of the problem at hand.

\begin{equation}
\label{eq:loss}
\text{loss} = \text{MSE}(z) + \text{MSE}(y)
\end{equation}

Adding to this, the use of MSE in this loss function serves multiple purposes. MSE is a widely adopted choice in machine learning for regression tasks due to its mathematical properties, such as convexity, differentiability. More specifically, the key factor lies in differentiability. The straightforward calculation of gradients is allowed, which is essential for optimizing deep neural networks.

\begin{figure*}[t!]
  \hskip 2cm % or \quad or adjust the value as needed
  \centering
  \includegraphics[width=0.85\textwidth]{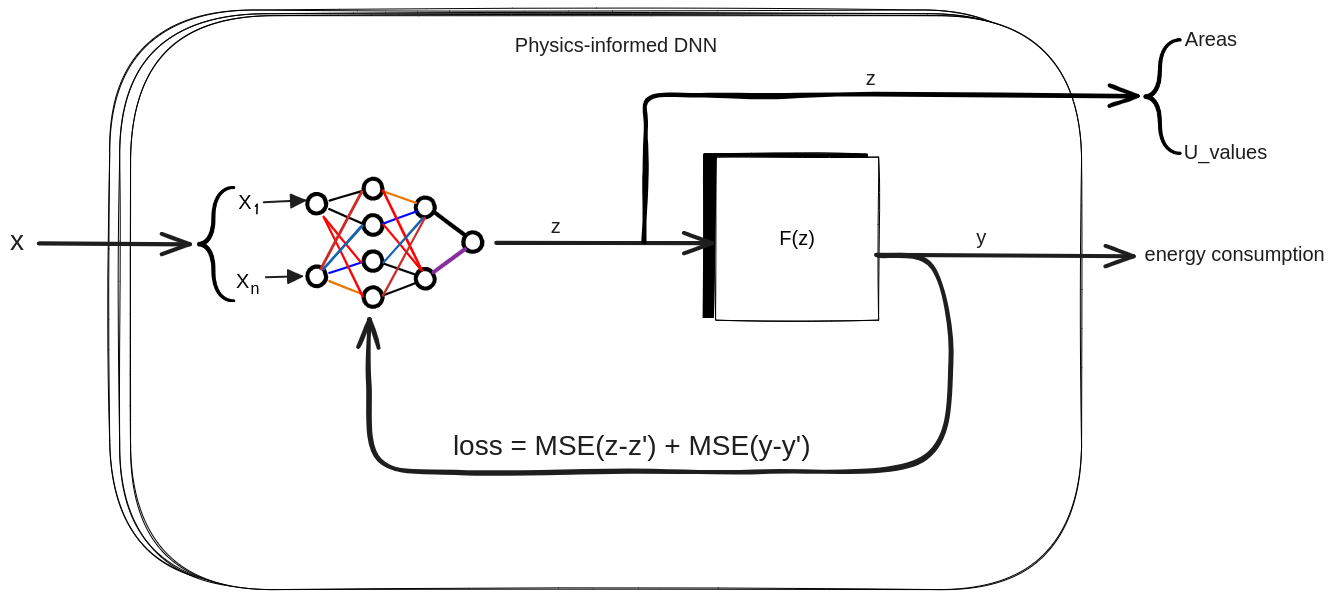}
  \caption{Physics-informed Deep Neural Network}
  \label{fig:DNN}
\end{figure*}

\subsection{Energy consumption calculations}
\label{subsec:calculations}
Hereby, we present a step-by-step description of the calculations involved in estimating the annual energy consumption of a building. These calculations can be seen in Figure \ref{fig:DNN} and are represented as the Physics-informed function $F_{}$. The computations take into account various building characteristics and environmental factors, ultimately providing insights into the energy requirements. The key variables utilized in the calculations are categorized as follows:

\begin{itemize}
    \item \textbf{Building Area Components:} These encompass the areas associated with different structural elements of the building, specifically, the basement/slab, roof/attic, walls, doors, and windows. These areas are denoted as $A_{\text{Basement/slab}}$, $A_{\text{Roof/attic}}$, $A_{\text{Walls}}$, $A_{\text{Doors}}$, and $A_{\text{Windows}}$, respectively.

\vspace{\baselineskip}

    \item \textbf{Thermal Transmittance (U-values):} The U-values signify the rate of heat transfer through building components, including $U_{\text{Basement/slab}}$, $U_{\text{Roof/attic}}$, $U_{\text{Walls}}$, $U_{\text{Doors}}$, and $U_{\text{Windows}}$. These values are indicative of the thermal insulation properties of the materials used.

\vspace{\baselineskip}

    \item \textbf{Ventilation and Heat Gain Parameters:} The model incorporates two vital parameters: the air exchange rate ($h$) and the specific heat gains ($Q$). The air exchange rate ($h$) reflects the rate at which outdoor air enters and exits the building, influencing heat loss or gain. Specific heat gains ($Q$) account for internal heat sources within the building.
\end{itemize}

\subsubsection{Envelope heat losses}

The heat losses through individual building envelope components (i.e., basement/slab, roof/attic, walls, doors, and windows) are calculated using the following equation:

\begin{equation}
\text{Heat Loss}_{\text{Envelope}} = \sum_{i=1}^{5} A_i \cdot U_i \cdot \Delta T \cdot \frac{192 \cdot 24}{1000}
\end{equation}

Where:
\begin{align*}
    A_i & : \text{Area of component } i \\
    U_i & : \text{U-value of component } i \\
    \Delta T & : \text{Temperature difference (18.9°C)}
\end{align*}

\subsubsection{Thermal bridges}

Thermal bridges denote localized areas with increased heat loss. In this calculation, thermal bridges are considered to account for 3\% of the total envelope heat losses:

\begin{equation}
\text{Thermal Bridges} = 0.03 \cdot \text{Heat Loss}_{\text{Envelope}}
\end{equation}

\subsubsection{Ventilation heat losses}

The heat losses due to ventilation are quantified by the following equation:

\begin{equation}
\text{Heat Loss}_{\text{Ventilation}} = V \cdot h \cdot 0.34 \cdot \Delta T \cdot \frac{192 \cdot 24}{1000}
\end{equation}

Where:
\begin{align*}
    V & : \text{Useful area} \\
    h & : \text{Air exchange rate}
\end{align*}

\subsubsection{Total heat losses}

The total heat losses are obtained by summing the envelope heat losses, thermal bridges, and ventilation heat losses:

\begin{equation}
\begin{aligned}
\text{Heat Loss}_{\text{Total}} &= \text{Heat Loss}_{\text{Envelope}} \\
&\quad + \text{Thermal Bridges} + \text{Heat Loss}_{\text{Ventilation}}
\end{aligned}
\end{equation}

\subsubsection{Total heat gains}

The heat gains attributed to specific heat gains within the building are calculated as follows:

\begin{equation}
\text{Heat Gains}_{\text{Total}} = Q \cdot V
\end{equation}

\subsubsection{Energy consumption}

The final energy consumption estimate is determined as the difference between total heat losses and total heat gains, with adjustments made based on a factor that considers the heat gain usage factor, which is influenced by the building type. The equation for energy consumption is:

\begin{equation}
\begin{aligned}
\text{Energy Consumption} &= \text{Heat Loss}_{\text{Total}} \\
&\quad - \text{Heat Gains}_{\text{Total}}  \cdot \text{HGUF}
\end{aligned}
\end{equation}

Where:

\begin{align*}
    HGUF & : \text{Heat Gain Usage Factor}
\end{align*}
\begin{equation}
\begin{aligned}
\text{Heat Gain Usage Factor} &= \frac{1 - \text{constant\_1} }{1 - \text{constant\_2}}
\end{aligned}
\end{equation}

Where: 

\begin{equation}
\begin{aligned}
\text{constant\_1} &= \left(\frac{\text{Heat Gains}_{\text{Total}}}{\text{Heat Loss}_{\text{Total}}}\right)^{\text{Building Time Constant}}
\end{aligned}
\end{equation}

\begin{equation}
\begin{aligned}
\text{constant\_2} &= \left(\frac{\text{Heat Gains}_{\text{Total}}}{\text{Heat Loss}_{\text{Total}}}\right)^{\text{Building Time Constant} + 1}
\end{aligned}
\end{equation}

Note here that, the Building Time Constant is determined based on the building type and thermal parameters.

In summary, these mathematical equations integrate building characteristics, thermal properties, and environmental conditions to yield an accurate estimate of a building's energy consumption. This estimate plays a crucial role in virtual energy audits and informs energy-efficient building design.

\section{Experiments}

\label{sec:experiments}

\subsection{Experimental Setup}

We use a 10-fold cross-validation procedure to train and
test the proposed model.The dataset is partitioned into training and testing sets using an 80-20 split, where 80\% of the data is allocated to the training set and the remaining 20\% to the testing set. Following this, the training set is further divided into a training subset and a validation subset using a 15-85 split, with 15\% of the original dataset dedicated to the training subset and the remaining 85\% assigned to the validation subset. Furthermore, a unique MinMax scaler for each one of the datasets, input, output and validation are used. We employed the Adam optimizer with a learning rate of 0.001 for the training of our neural network model. Additionally, a learning rate scheduler, specifically the ReduceLROnPlateau scheduler, was utilized to dynamically adjust the learning rate during training. The scheduler reduced the learning rate by a factor of 0.1 after a period of 5 epochs without improvement in validation loss. The selection of the learning rate and the use of a learning rate scheduler were motivated by their established effectiveness in optimizing neural network models. The optimization process was monitored based on the validation loss, with the goal of enhancing the model's generalization performance. Also, we apply early stopping and stop training, if the validation loss has stopped decreasing for eight consecutive epochs.The model is created using the PyTorch Lightning library \cite{FalconPyTorchLightning2019}. All experimental procedures were conducted on a Dell Vostro 15 3000 series laptop with 8 cores and 16Gb of RAM.

\subsection{Evaluation Metrics}
\label{sub:metrics}
 This subsection introduces three widely used evaluation metrics, R-squared ($R^2$),Root Mean Squared Error (RMSE) and Normalized Root Mean Squared Error (NRMSE), along with their mathematical equations that are used to evaluate our model. They cover the goodness of fit, the error magnitude and scale. It needs to be mentioned that NRMSE is defined as:

 \begin{equation}
NRMSE = \frac{RMSE}{Y_{\text{max}} - Y_{\text{min}}}
\end{equation}
Where:
\begin{align*}
Y_{\text{max}} & \text{ is the maximum value of the dependent variable.} \\
Y_{\text{min}} & \text{ is the minimum value of the dependent variable.}
\end{align*}

\subsection{Results}

\begin{table*}[!th]
\centering
\caption{Model evaluation results for output variables}
\label{tab:results}
\begin{tabular}{p{2.5cm}p{2.5cm}p{2.5cm}p{2.5cm}} % Remove the "MAPE" column
\hline
\textbf{Variable} & \textbf{R$^2$} & \textbf{RMSE} & \textbf{NRMSE} \\ % Remove "MAPE" here
\hline
area\_Basement/slab & 0.67 $\pm$ 0.04 & 219.61 $\pm$ 7.88 & 0.11 $\pm$ 0.00 \\
area\_Roof/attic & 0.75 $\pm$ 0.04 & 189.35 $\pm$ 8.81 & 0.08 $\pm$ 0.01 \\
area\_Walls & 0.69 $\pm$ 0.03 & 736.26 $\pm$ 32.55 & 0.13 $\pm$ 0.01 \\
area\_doors & 0.76 $\pm$ 0.03 & 9.95 $\pm$ 0.40 & 0.10 $\pm$ 0.01 \\
area\_windows & 0.95 $\pm$ 0.01 & 100.85 $\pm$ 6.94 & 0.04 $\pm$ 0.00 \\
U\_Basement/slab & 0.29 $\pm$ 0.07 & 0.20 $\pm$ 0.01 & 0.17 $\pm$ 0.01 \\
U\_Roof/attic & 0.05 $\pm$ 0.09 & 0.31 $\pm$ 0.01 & 0.17 $\pm$ 0.01 \\
U\_Walls & 0.16 $\pm$ 0.06 & 0.24 $\pm$ 0.01 & 0.16 $\pm$ 0.01 \\
U\_doors & 0.04 $\pm$ 0.05 & 0.65 $\pm$ 0.01 & 0.26 $\pm$ 0.01 \\
U\_windows & 0.09 $\pm$ 0.04 & 0.24 $\pm$ 0.00 & 0.23 $\pm$ 0.01 \\
air\_exchange\_rate & 0.41 $\pm$ 0.04 & 0.08 $\pm$ 0.02 & 0.01 $\pm$ 0.00 \\
specific\_heat\_gains & 0.21 $\pm$ 0.06 & 9.96 $\pm$ 0.37 & 0.24 $\pm$ 0.01 \\
\hline
\end{tabular}
\end{table*}

The results of the proposed model before are reported in Tables \ref{tab:results} and \ref{tab:single_value_results}. As stated in the previous subsection we evaluate the model's performance with $R^2$, RMSE and NRMSE. The results are presented in the form of mean values accompanied by $\pm$ standard deviations. The tables provide a snapshot of the performance of the model across a diverse set of variables, demonstrating the range of prediction accuracy and goodness-of-fit within the dataset.

\begin{table}[b]
\centering
\caption{Model evaluation results for Energy Consumption}
\label{tab:single_value_results}
\begin{tabular}{|l|c|c|c|c|} % Add a new column for NRMSE
\hline
\textbf{Variable} & \textbf{R$^2$} & \textbf{RMSE} & \textbf{NRMSE} \\
\hline
Energy Consumption & 0.87 $\pm$ 0.01 & 100.79 $\pm$ 7.82 & 0.065 $\pm$ 0.01 \\
\hline
\end{tabular}
\end{table}

More specifically, Table \ref{tab:results} lists the results of the DNN for the multi-output regression problem. While some variables show average results, like area\_Walls and area\_Basement/slab with reasonable $R^2$, RMSE and NRMSE values the majority of the variables exhibit substantial variation. Note here that the presence extreme maximum and minimum values for all the metrics underscore the heterogeneity in the model's performance. More specifically, while some variables indicate high accuracy with low NRMSE and RMSE as well as high R$^2$ values (e.g., area\_windows), others demonstrate substantial prediction deviations and weaker fits (e.g. U\_Roof/attic and U\_windows). These disparities are instrumental in identifying specific areas where the model excels and where improvements are be required. Note also that such deviations are expected due to the fact that we have only five input features in contrast to the twelve predicted ones. Finally, the large differences in RMSE values can be attributed to the difference in the order of the magnitude of the variables as also confirmed by the relatively close values with respect to NRMSE.

Subsequently, Table \ref{tab:single_value_results} highlights the model evaluation results regarding the energy consumption. Notably, R$^2$ for energy consumption is reported at \textbf{0.87 $\pm$ 0.01}, RMSE at \textbf{102.69 $\pm$ 7.82}, while NRMSE stands at \textbf{0.065 $\pm$ 0.01}. Note here that energy consumption is a multifaceted variable influenced, in turn, by various other predicted variables (contained in vector z) which, as aforementioned, are not always perfectly accurate. However, the results here are satisfying with relatively low errors as demonstrated by the NRMSE. Ultimately, note that the value of the custom loss function is \textbf{0.60 $\pm$ 0.03} at the end of the training cycle for the ten folds.

\section{Limitations}
\label{sec:lmitations}

Our study represents a substantial step toward energy consumption prediction based on envelope component's heat losses and has yielded promising results; however, it is essential to recognize several key limitations that warrant consideration. One of the primary constraints of our study lies in the availability of data. Like all ML and DL solutions, this problem requires extensive datasets encompassing a diversity of building-related variables. These datasets should ideally include data from a multitude of buildings to effectively address the challenges at hand. Despite our efforts to collect and utilize data from diverse sources, the dataset is quite small for such a challenging task. Another challenge can be identified in what concerns the quality of the data. Given the notable presence of outliers and potential erroneous measurements within our datasets, and as their removal was prohibitive due to dataset size limitations, the resulting training and testing subsets exhibited several inconsistencies that are speculated to affect the regression results.
Despite the aforementioned limitations, the contribution of the present study is crucial for the energy efficiency domain, as it provides an innovative methodology for data-driven, decomposed energy performance prediction, introducing a unique physics-informed DNN, and supplying valuable insights for renovation endeavors.

\section{Conclusions \& Future Work}
\label{sec:conclusions}

In this paper, we presented a novel approach that couples physics and DL methodologies to estimate the energy performance of residential buildings decomposed across the individual components of their envelope (e.g. windows, doors, floors, roofs, basements). The current study is anchored in an exploration of distinctive datasets containing comprehensive information on the structural, envelope, and energy characteristics of actual buildings located in Riga, Latvia. In this context, we introduced an innovative enhanced loss function that integrates linear physics equations and is designed to enhance our ability to predict the properties of the individual envelope segments of residential buildings alongside their energy consumption, primarily driven by the heat losses associated with these segments. With respect to the decomposed losses per envelope component the average relative error (NRMSE) ranges around 0.14 with an average goodness-of-fit (R$^2$) of 0.42, yielding a rather moderate performance. However, regarding the total energy consumption the results are significantly improved, with an NRMSE of 0.065 and an R$^2$ of 0.87, highlighting that the proposed methodology can provide accurate building total energy consumption predictions. Despite the specific accuracy levels, it should be noted that this piece of work lays the methodological foundations for the transformation of the energy efficiency sector through an automated, data-driven methodology that can significantly accelerate the energy audit procedure within residential buildings. Additionally, the proposed methodology can be used by several stakeholders of the building energy efficiency sector (e.g. building owners, energy auditors, and policymakers) towards renovation and retrofit initiatives, or even recommendation systems for the maximization of the impact of energy efficiency renovation measures.

As we move forward, further research and development in this area can potentially refine and expand upon our approach. In this context, we propose the mass concentration and digitalization of energy audits aiming at larger scale datasets that can improve the learning potential of DNN models for energy efficiency performance estimation. Additionally, we anticipate the integration of more physics models and that the incorporation of additional variables related to building performance will continue to improve the accuracy of our predictions. Ultimately, another potential step forward, would be the development of a user-friendly web application that would render our methodology accessible to a wider audience, including building owners, energy auditors, and policymakers.

\section*{Acknowledgment}

This work has been funded by the European Union’s Horizon 2020 research and innovation program under the I-NERGY project, grant agreement No. 101016508. The content of the paper is the sole responsibility of its authors and does not necessary reflect the views of the EC.

\bibliographystyle{IEEEtran}
\bibliography{references}

\begin{thebibliography}{10}
\providecommand{\url}[1]{#1}
\csname url@samestyle\endcsname
\providecommand{\newblock}{\relax}
\providecommand{\bibinfo}[2]{#2}
\providecommand{\BIBentrySTDinterwordspacing}{\spaceskip=0pt\relax}
\providecommand{\BIBentryALTinterwordstretchfactor}{4}
\providecommand{\BIBentryALTinterwordspacing}{\spaceskip=\fontdimen2\font plus
\BIBentryALTinterwordstretchfactor\fontdimen3\font minus \fontdimen4\font\relax}
\providecommand{\BIBforeignlanguage}[2]{{%
\expandafter\ifx\csname l@#1\endcsname\relax
\typeout{** WARNING: IEEEtran.bst: No hyphenation pattern has been}%
\typeout{** loaded for the language `#1'. Using the pattern for}%
\typeout{** the default language instead.}%
\else
\language=\csname l@#1\endcsname
\fi
#2}}
\providecommand{\BIBdecl}{\relax}
\BIBdecl

\bibitem{sarmas2022meta}
E.~Sarmas, E.~Spiliotis, V.~Marinakis, T.~Koutselis, and H.~Doukas, ``A meta-learning classification model for supporting decisions on energy efficiency investments,'' \emph{Energy and Buildings}, vol. 258, p. 111836, 2022.

\bibitem{Terhaar2022}
\BIBentryALTinterwordspacing
{Terhaar}, {Fr{\"o}licher}, {Aschwanden}, {Friedlingstein}, and {Joos}, ``Adaptive emission reduction approach to reach any global warming target,'' \emph{Nature Climate Change}, vol.~12, no.~12, pp. 1136--1142, Dec 2022. [Online]. Available: \url{https://doi.org/10.1038/s41558-022-01537-9}
\BIBentrySTDinterwordspacing

\bibitem{irena2015renewable}
R.~Irena, ``Renewable energy target setting,'' \emph{International Renewable Energy Agency: Abu Dhabi, United Arab Emirates}, 2015.

\bibitem{Karakolis202261}
\BIBentryALTinterwordspacing
E.~Karakolis, S.~Pelekis, S.~Mouzakitis, O.~Markaki, K.~Papapostolou, G.~Korbakis, and J.~Psarras, ``{ARTIFICIAL INTELLIGENCE FOR NEXT GENERATION ENERGY SERVICES ACROSS EUROPE - THE I-NERGY PROJECT},'' 2022, Conference paper, p. 61 – 68. [Online]. Available: \url{https://www.scopus.com/inward/record.uri?eid=2-s2.0-85137606268\&partnerID=40\&md5=29734e59b14194f1ea6ee4c070f17422}
\BIBentrySTDinterwordspacing

\bibitem{Wehrmeister2022TheApplications}
K.~A. Wehrmeister, E.~Bothos, V.~Marinakis, B.~Magoutas, A.~Pastor, L.~Carreras, and A.~Monti, ``{The BD4NRG Reference Architecture for Big Data Driven Energy Applications},'' \emph{13th International Conference on Information, Intelligence, Systems and Applications, IISA 2022}, 2022.

\bibitem{Pau2022MATRYCSDomain}
M.~Pau, P.~Kapsalis, Z.~Pan, G.~Korbakis, D.~Pellegrino, and A.~Monti, ``{MATRYCS A Big Data Architecture for Advanced Services in the Building Domain},'' \emph{Energies}, vol.~15, no.~7, p. 2568, 4 2022.

\bibitem{Pelekis2022InPerformance}
S.~Pelekis, E.~Karakolis, F.~Silva, V.~Schoinas, S.~Mouzakitis, G.~Kormpakis, N.~Amaro, and J.~Psarras, ``{In Search of Deep Learning Architectures for Load Forecasting: A Comparative Analysis and the Impact of the Covid-19 Pandemic on Model Performance},'' in \emph{IISA 2022 - 13th International Conference on Information, Intelligence, Systems and Applications (IISA)}.\hskip 1em plus 0.5em minus 0.4em\relax IEEE, 7 2022, pp. 1--8.

\bibitem{Pelekis2023ADrivers}
\BIBentryALTinterwordspacing
S.~Pelekis, I.-K. Seisopoulos, E.~Spiliotis, T.~Pountridis, E.~Karakolis, S.~Mouzakitis, and D.~Askounis, ``{A comparative assessment of deep learning models for day-ahead load forecasting: Investigating key accuracy drivers},'' \emph{Sustainable Energy, Grids and Networks}, vol.~36, p. 101171, 12 2023. [Online]. Available: \url{https://linkinghub.elsevier.com/retrieve/pii/S2352467723001790}
\BIBentrySTDinterwordspacing

\bibitem{Pelekis2023DeepTSF:Forecasting}
\BIBentryALTinterwordspacing
S.~Pelekis, E.~Karakolis, T.~Pountridis, G.~Kormpakis, G.~Lampropoulos, S.~Mouzakits, and D.~Askounis, ``{DeepTSF: Codeless machine learning operations for time series forecasting},'' \emph{arXiv}, 7 2023. [Online]. Available: \url{https://arxiv.org/abs/2308.00709}
\BIBentrySTDinterwordspacing

\bibitem{Pelekis2023TargetedTechniques}
\BIBentryALTinterwordspacing
S.~Pelekis, A.~Pipergias, E.~Karakolis, S.~Mouzakitis, F.~Santori, M.~Ghoreishi, and D.~Askounis, ``{Targeted demand response for flexible energy communities using clustering techniques},'' \emph{Sustainable Energy, Grids and Networks}, vol.~36, p. 101134, 12 2023. [Online]. Available: \url{https://linkinghub.elsevier.com/retrieve/pii/S235246772300142X}
\BIBentrySTDinterwordspacing

\bibitem{michalakopoulos2024machine}
V.~Michalakopoulos, E.~Sarmas, I.~Papias, P.~Skaloumpakas, V.~Marinakis, and H.~Doukas, ``A machine learning-based framework for clustering residential electricity load profiles to enhance demand response programs,'' \emph{Applied Energy}, vol. 361, p. 122943, 2024.

\bibitem{Pelekis2023}
\BIBentryALTinterwordspacing
S.~Pelekis, E.~Sarmas, A.~Georgiadou, E.~Karakolis, C.~Ntanos, N.~Dimitropoulos, G.~Kormpakis, and H.~Doukas, ``Twinp2g: A digital twin architecture for optimal power-to-gas planning,'' 2023. [Online]. Available: \url{https://www.esociety-conf.org/wp-content/uploads/2023/03/2\_ES2023\_F\_107\_Pelekis.pdf}
\BIBentrySTDinterwordspacing

\bibitem{KOUTSANDREAS2023101233}
\BIBentryALTinterwordspacing
D.~Koutsandreas, G.~P. Trachanas, I.~Pappis, A.~Nikas, H.~Doukas, and J.~Psarras, ``{A multicriteria modeling approach for evaluating power generation scenarios under uncertainty: The case of green hydrogen in Greece},'' \emph{Energy Strategy Reviews}, vol.~50, p. 101233, 2023. [Online]. Available: \url{https://www.sciencedirect.com/science/article/pii/S2211467X23001839}
\BIBentrySTDinterwordspacing

\bibitem{FRILINGOU2023102934}
\BIBentryALTinterwordspacing
N.~Frilingou, G.~Xexakis, K.~Koasidis, A.~Nikas, L.~Campagnolo, E.~Delpiazzo, A.~Chiodi, M.~Gargiulo, B.~McWilliams, T.~Koutsellis, and H.~Doukas, ``Navigating through an energy crisis: Challenges and progress towards electricity decarbonisation, reliability, and affordability in italy,'' \emph{Energy Research and Social Science}, vol.~96, p. 102934, 2023. [Online]. Available: \url{https://www.sciencedirect.com/science/article/pii/S2214629622004376}
\BIBentrySTDinterwordspacing

\bibitem{van2023multimodel}
D.-J. van~de Ven, S.~Mittal, A.~Gambhir, R.~D. Lamboll, H.~Doukas, S.~Giarola, A.~Hawkes, K.~Koasidis, A.~C. K{\"o}berle, H.~McJeon \emph{et~al.}, ``A multimodel analysis of post-glasgow climate targets and feasibility challenges,'' \emph{Nature Climate Change}, pp. 1--9, 2023.

\bibitem{KOASIDIS2023205}
\BIBentryALTinterwordspacing
K.~Koasidis, A.~Nikas, and H.~Doukas, ``Why integrated assessment models alone are insufficient to navigate us through the polycrisis,'' \emph{One Earth}, vol.~6, no.~3, pp. 205--209, 2023. [Online]. Available: \url{https://www.sciencedirect.com/science/article/pii/S2590332223000891}
\BIBentrySTDinterwordspacing

\bibitem{lasvaux2010study}
S.~Lasvaux, ``Study of a simplified model for the life cycle assessment of buildings,'' Ph.D. dissertation, {\'E}cole Nationale Sup{\'e}rieure des Mines de Paris, 2010.

\bibitem{stafford2012building}
A.~Stafford, M.~Bell, and C.~Gorse, ``Building confidence--a working paper,'' \emph{Centre for Low Carbon Futures: York, UK}, 2012.

\bibitem{AHMAD2018301}
\BIBentryALTinterwordspacing
T.~Ahmad, H.~Chen, Y.~Guo, and J.~Wang, ``A comprehensive overview on the data driven and large scale based approaches for forecasting of building energy demand: A review,'' \emph{Energy and Buildings}, vol. 165, pp. 301--320, 2018. [Online]. Available: \url{https://www.sciencedirect.com/science/article/pii/S0378778817329225}
\BIBentrySTDinterwordspacing

\bibitem{HAIDAR2023109806}
\BIBentryALTinterwordspacing
N.~Haidar, N.~Tamani, Y.~Ghamri-Doudane, and A.~Boujou, ``Selective reinforcement graph mining approach for smart building energy and occupant comfort optimization,'' \emph{Building and Environment}, vol. 228, p. 109806, 2023. [Online]. Available: \url{https://www.sciencedirect.com/science/article/pii/S0360132322010368}
\BIBentrySTDinterwordspacing

\bibitem{HOMOD2023105689}
\BIBentryALTinterwordspacing
R.~Z. Homod, Z.~M. Yaseen, A.~K. Hussein, A.~Almusaed, O.~A. Alawi, M.~W. Falah, A.~H. Abdelrazek, W.~Ahmed, and M.~Eltaweel, ``Deep clustering of cooperative multi-agent reinforcement learning to optimize multi chiller hvac systems for smart buildings energy management,'' \emph{Journal of Building Engineering}, vol.~65, p. 105689, 2023. [Online]. Available: \url{https://www.sciencedirect.com/science/article/pii/S2352710222016953}
\BIBentrySTDinterwordspacing

\bibitem{LONG2023480}
\BIBentryALTinterwordspacing
L.~D. Long, ``An ai-driven model for predicting and optimizing energy-efficient building envelopes,'' \emph{Alexandria Engineering Journal}, vol.~79, pp. 480--501, 2023. [Online]. Available: \url{https://www.sciencedirect.com/science/article/pii/S1110016823007251}
\BIBentrySTDinterwordspacing

\bibitem{KALOGIROU2004383}
\BIBentryALTinterwordspacing
S.~A. Kalogirou, ``Optimization of solar systems using artificial neural-networks and genetic algorithms,'' \emph{Applied Energy}, vol.~77, no.~4, pp. 383--405, 2004. [Online]. Available: \url{https://www.sciencedirect.com/science/article/pii/S0306261903001533}
\BIBentrySTDinterwordspacing

\bibitem{BORGEDIEZ2015821}
\BIBentryALTinterwordspacing
D.~Borge-Diez, A.~Colmenar-Santos, C.~Pérez-Molina, and África López-Rey, ``Geothermal source heat pumps under energy services companies finance scheme to increase energy efficiency and production in stockbreeding facilities,'' \emph{Energy}, vol.~88, pp. 821--836, 2015. [Online]. Available: \url{https://www.sciencedirect.com/science/article/pii/S0360544215008981}
\BIBentrySTDinterwordspacing

\bibitem{AMASYALI20181192}
\BIBentryALTinterwordspacing
K.~Amasyali and N.~M. El-Gohary, ``A review of data-driven building energy consumption prediction studies,'' \emph{Renewable and Sustainable Energy Reviews}, vol.~81, pp. 1192--1205, 2018. [Online]. Available: \url{https://www.sciencedirect.com/science/article/pii/S1364032117306093}
\BIBentrySTDinterwordspacing

\bibitem{srid}
\BIBentryALTinterwordspacing
Spatial reference identifier (srid). [Online]. [Online]. Available: \url{https://desktop.arcgis.com/en/arcmap/latest/manage-data/using-sql-with-gdbs/what-is-an-srid.html}
\BIBentrySTDinterwordspacing

\bibitem{postgis2018postgis}
\BIBentryALTinterwordspacing
P.~P.~S. Committee \emph{et~al.}, ``Postgis, spatial and geographic objects for postgresql,'' 2018. [Online]. Available: \url{https://postgis.net}
\BIBentrySTDinterwordspacing

\bibitem{psqldocs}
\BIBentryALTinterwordspacing
P.~D. Team, ``{PostgreSQL Documentation}.'' [Online]. Available: \url{https://www.postgresql.org/docs/}
\BIBentrySTDinterwordspacing

\bibitem{thermaltransmittance}
J.~Willoughby, ``30 - insulation,'' in \emph{Plant Engineer's Reference Book (Second Edition)}, second edition~ed., D.~A. Snow, Ed.\hskip 1em plus 0.5em minus 0.4em\relax Oxford: Butterworth-Heinemann, 2002, pp. 30--1--30--18.

\bibitem{FalconPyTorchLightning2019}
\BIBentryALTinterwordspacing
W.~Falcon and {The PyTorch Lightning team}, ``{PyTorch Lightning},'' Mar. 2019. [Online]. Available: \url{https://github.com/Lightning-AI/lightning}
\BIBentrySTDinterwordspacing

\end{thebibliography}

\end{document}